# Activity Recognition Using A Combination of Category Components And Local Models for Video Surveillance

Weiyao Lin, Ming-Ting Sun, Radha Poovendran, and Zhengyou Zhang

*Abstract*—This paper presents a novel approach for automatic recognition of human activities for video surveillance applications. We propose to represent an activity by a combination of category components, and demonstrate that this approach offers flexibility to add new activities to the system and an ability to deal with the problem of building models for activities lacking training data. For improving the recognition accuracy, a Confident-Frame- based Recognition algorithm is also proposed, where the video frames with high confidence for recognizing an activity are used as a specialized local model to help classify the remainder of the video frames. Experimental results show the effectiveness of the proposed approach.

*Index Terms*—Event Detection, Category Components, Local Model, Video Surveillance

## 1. INTRODUCTION AND RELATED WORK

VIDEO surveillance is of increasing importance in many applications, including elder care, home nursing, and unusual event alarming [1-4]. Automatic activity recognition plays a key part in video surveillance. In this paper, we focus on addressing the following three key issues for event recognition.

### A. The Flexibility of the System for Adding New Events

In many applications, people may often want to add new events of interest into the recognition system. It is desirable that the existing models in the system are not affected or do not need to be re-constructed when new events are added. Using most existing activity recognition algorithms [12-17][26-27], the whole system has to be re-trained or re-constructed for the new added events. Some methods [5,6,24] try to use a similarity metric so that different events can be clustered into different groups. This approach has more flexibility for new added events. However, due to the uncertain nature of the activity instances, it is difficult to find a suitable feature set for all samples of an event to be clustered closely around a center.

### B. Recognition of Events Lacking Training Samples

In many surveillance applications, events of interest may only occur rarely (e.g., most unusual events such as a heart attack or falling down stairs). For these events, it is difficult to collect sufficient training samples for learning the unusual event models. In this case, many event detection algorithms [8-9] which require large numbers of training data become unsuitable. Methods for learning from small numbers of examples are needed [5-7][26-29]. In this paper, we call these Lacking-Training-Sample (LTS) events.

Several algorithms have been proposed to address the difficulty of recognizing LTS events. Wu et al. [27] and Amari et al. [26] try to solve the unbalanced-training-data problem by using a conformal transform to adapt the Support Vector Machine (SVM) kernels. However, these methods still need boundary training samples (samples around class boundaries) to obtain good support vectors for differentiating different classes, while in reality the insufficient training set may not include these boundary training samples. Other researchers [28,29] try to improve the estimation of model parameters (e.g. the Gaussian covariance matrix) for cases of limited training samples. However, these methods do not work well if the limited training data are not sufficient to fully represent events.

### C. Accuracy for Recognizing Human Activities

Recognition accuracy is always a major concern for automatic event recognition. Many algorithms have been proposed which try to detect human activities with high accuracy. Cristani et al. [22], Zhang et al. [35], and Dupant et al. [36] focus on developing suitable multi-stream fusion methods to combine features from different streams (e.g., audio and video) to improve the recognition accuracy. Cristani et al. [22] propose an AVC matrix for audio and video stream fusion. Dupant et al. [36] propose to use Weighted Multiplication for combining multi-stream data. Zhang et al. [35] compare different stream-combining methods such as Weighted Multiplication and Early Integration. Models such as HMM (Hidden Markov Model) or DBN (Dynamic Bayesian network) [12-14], state machine [4,15], Adaboost [16-17], and SVM [26,27] are widely used in these works for activity recognition. However, most of these methods only work well in their assumed scenarios and have limitations or lower accuracy if applied to other scenarios. Therefore, it is always desirable to develop new algorithms to improve the recognition accuracy.

The contribution of this paper is summarized as follows: (1) To address the flexibility problem for adding new events, we propose to use a Category Feature Vector (CFV) based model to represent an event. (2) To address the problem of recognizing events which lack training samples (LTS events), we propose a new approach to derive models for the LTS events from the parts from other trained related events. (3) To address the accuracy problem for recognition algorithms, we propose a

Weiyao Lin, Ming-Ting Sun and Radha Poovendran are with the Department of Electrical Engineering, University of Washington, Seattle, USA (e-mail: {wylin,mts,rp3}@u.washington.edu).
Zhengyou Zhang is with the Microsoft Research, Microsoft Corp., Redmond, USA (e-mail: zhang@microsoft.com).

Confident-Frame-based Recognition algorithm (CFR) to improve the accuracy of the recognition.

The rest of the paper is organized as follows. Section 2 describes our approach to represent activities. Based on this activity representation, Section 3 discusses the flexibility of our method for training new activities. Section 4 describes our proposed method to train models for events lacking training data. In Section 5, we present our Confident-Frame-based Recognition algorithm (CFR) to improve the recognition accuracy. Experimental results are shown in Section 6. We conclude the paper in Section 7.

## 2. ACTIVITY REPRESENTATION

For flexible classification, activities can be described by a combination of feature attributes. For example, a set of human activities (*Inactive*, *Active*, *Walking*, *Running*, *Fighting*) [11] can be differentiated using a combination of attributes of two features: *Change of Body Size* (CBS) and *Speed*. Each feature can have attributes *High*, *Medium*, and *Low*. *Inactive*, which represents a static person, can be described as *Low CBS* and *Low Speed*. *Active*, which represents a person making movements but little translations, can be described as *Medium CBS* and *Low Speed*. *Walking*, representing a person making movements and translations, can be described as *Medium CBS* and *Medium Speed*. *Running*, which is similar to *Walking* but with a larger translation, can be described as *High CBS* and *High Speed*. *Fighting*, which has large movements with small translations, can be described as *High CBS* and *Low Speed*.

It is efficient to represent the activities by the combination of *Feature-Attributes* as shown in the above example. A relatively small number of features and attributes can describe and differentiate a large number of activities ($n$ features with $m$ attributes could describe $m^n$ activities). However, this approach has low robustness. The misclassification of one feature attribute can easily lead to a completely wrong result. Furthermore, the extent of "*Medium*", "*Low*", or "*High*" is difficult to define.

The above *Feature-Attribute* description for representing activities can be extended to a description by a combination of Category Features Vectors (CFVs) with each CFV $F_i$ contains a set of correlated features. $F_i$ is defined by $F_i = [f_{1i}, f_{2i}, ... f_{mi}]^T$, where $f_{1i}, f_{2i}, ... f_{mi}$ are correlated features related to the *same* category $i$. Each CFV can be further represented by different models. For example, using a Gaussian Mixture Model (GMM) [11,31,32] as shown in Fig. 1, the likelihood function $p(F_i(t)|A_k)$ of the observed CFV $F_i(t)$ for video frame $t$, given activity $A_k$ can be described as

$$p(F_i(t) | A_k) \approx \sum_{j=1}^{M_{Fi,Ak}} \pi_j^{Fi,Ak} \cdot N(\mu_j^{Fi,Ak}, \sigma_j^{Fi,Ak}) \quad (1)$$

where $\pi_j^{Fi,Ak}$ is the weight of the $j$-th Gaussian distribution $N(\mu_j^{Fi,Ak}, \sigma_j^{Fi,Ak})$ for the CFV of $F_i(t)$ given activity $A_k$. $\mu_j^{Fi,Ak}$ and $\sigma_j^{Fi,Ak}$ are the mean and variance for distribution $N(\mu_j^{Fi,Ak}, \sigma_j^{Fi,Ak})$, respectively. $\pi_j^{Fi,Ak}$ is normalized to make $p(F_i(t)|A_k)$ a proper probability distribution. $M_{Fi,Ak}$ is the number of Gaussian Mixtures for $F_i$ given $A_k$.

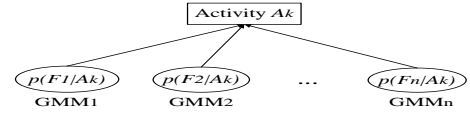

Fig. 1. Activity $A_k$ is described by a combination of GMMs with each GMM representing the distribution $p(F_i/A_k)$ of a Category Feature Vector $F_i$.

Essentially, CFV is the extension of the 'feature' in the *Feature-Attribute* description. Features with high correlations for describing activities are grouped into the same CFV. The GMM model $p(F_i(t)|A_k)$ is the extension of the 'feature attribute' in the *Feature-Attribute* description. With the use of the CFV representation, we will have more robustness in representing and recognizing activities compared to the *Feature-Attribute* description. It should be noted that although in this paper we use a GMM to represent a CFV, the proposed CFV representation is not limited to the GMM model. Other models such as HMM or DBN can also be used to represent a CFV.

In practice, CFVs can be formed by clustering features based on feature similarities such as *correlation coefficient* [30] or *K-L distance* [10][25]. In the experiments presented in this paper, the CFVs are formed by clustering the features based on their K-L distances. The similarity of two feature distributions can be approximated by the K-L distance in terms of the means and variances of the Gaussian distributions [10][25]:

$$D(f_i, f_j) = \sum_{k \in A} \left\{ (\mu_{i,k} - \mu_{j,k})^2 \cdot \left( \frac{1}{\sigma_{i,k}^2} + \frac{1}{\sigma_{j,k}^2} \right) + \frac{\sigma_{i,k}^2}{\sigma_{j,k}^2} + \frac{\sigma_{j,k}^2}{\sigma_{i,k}^2} \right\} \quad (2)$$

where $f_i, f_j$ are two features, and $\mu_{i,k}$ and $\sigma_{i,k}$ are the mean and the variance of the probability distribution $p(f_i/A_k)$. By grouping correlated features into a CFV, the correlations of the features can be captured by the GMM. Also, we can reduce the total number of GMM models, which can facilitate the succeeding classifier which is based on fusing the GMM results. Furthermore, by separating the complete set of features into CFVs, it facilitates the handling of new added activities and the training of models for LTS events as described below.

## 3. HANDLING NEW ADDED ACTIVITIES

When new activities are added to the system, the already defined features may not be enough to differentiate all activities, necessitating the adding of new features. With our CFV-based representation, we only need to define new categories for the added features (i.e., define new CFVs) and train new models for them (i.e., add a new GMM for each new CFV), while keeping the other CFV-GMMs of the already existing events unchanged. For example, in Fig. 2, the original system has two activities $A_1$ and $A_2$, each activity has $n$ CFV-based GMM models to represent it (grey circles in Fig. 2). When a new activity $A_3$ is added to the system, new features are needed to differentiate the three activities. We define a new CFV $F_{n+1}$ with a GMM model $p(F_{n+1}/A_k)$ for these new features and add it to all activities $A_k$. The flexibility of our representation is that we only need to train new models for the new event $A_3$ as well as the newly-added model $p(F_{n+1}/A_k)$ for the existing events $A_1$ and $A_2$ (white circles in the Bold rectangle in Fig. 2), while keeping all the existing models in the original system unchanged (grey circles in Fig. 2). In practice, the number of the trained activities could be much larger compared to the number of new added events. The flexibility offered by the CFV-based system enables the possibility of designing a programmable system which can incrementally grow, instead of needing to retrain the whole system when a new event needs to be added. In contrast to the above, the models of traditional methods will become increasingly complicated with the addition of new features.

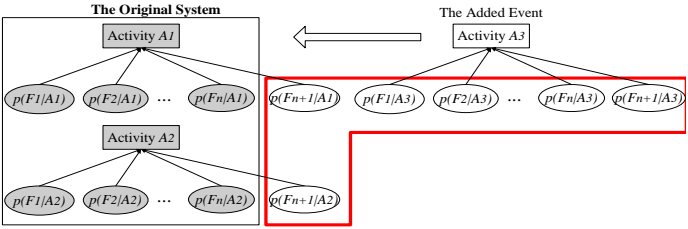

Fig. 2. The flexibility for adding a new event $A_3$. (Grey circles: models do not need to be changed, white circles: models need training)

## 4. TRAINING MODELS FOR LTS EVENTS

Since LTS events lack training data, we often do not have enough data to construct a representative model for these events. To solve this problem, we observe that people often describe a rare object by a combination of different parts from familiar objects. For example, people may describe a llama as an animal with a head similar to a camel and a body similar to a donkey. Similarly, with our CFV-based representation of activities, it is possible for us to derive a good initial LTS event model from the CFVs of the previously trained activities. For example, as shown in Fig. 3, we have trained two CFVs $F_{CBS}$ and $F_{Speed}$ for recognizing four events: *Active*, *Inactive*, *Walking*, and *Running*. $F_{CBS}$ is the CFV for the category *Change of Body Size* (*CBS*), and $F_{Speed}$ is the CFV for the category *Speed*. Assume *Fighting* is an event we try to recognize but lacking training data. For the CBS category, we can reason that the behavior of *Running* is the most similar among all the usual events to that of *Fighting*, therefore, the GMM for $F_{CBS}^{fighting}$ will be adapted from that of $F_{CBS}^{running}$. Similarly, for the Speed Category, we find that the behavior of *Active* is the most similar to that of *Fighting*, therefore, the GMM for $F_{Speed}^{fighting}$ will be adapted from that of $F_{Speed}^{active}$. In this way, we can have a good initial model for *Fighting* even if we lack training data.

We propose to generate models for LTS activities as follows. For each $CFV_i$ in category $i$ of the LTS activity $A_u$, find the trained GMM model ($GMM_i^{A_k}$) where the behavior of activity $A_k$ is most similar to the LTS activity $A_u$ in this *specific* category.

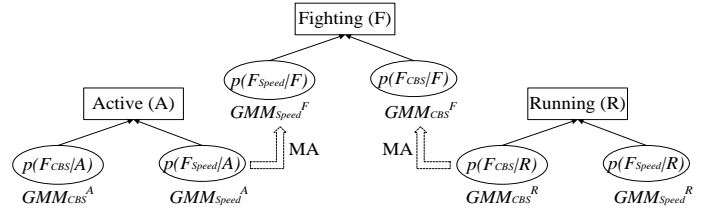

Fig. 3. The Training of an LTS Event *Fighting*.

This initial model can be adapted further to derive a new model $GMM_i^{A_u}$ using the limited training data and the MAP-based Adaptation (MA) [7][33]. MA is an extension of the EM algorithm which contains two steps:

(1) Update the parameters $\theta^{New}$ by the regular EM algorithm [34] with the limited training data.
(2) The GMM parameters are then adapted by the linear combination of the parameters $\theta^{New}$ and the parameters of the initial model $\theta^{Old}$ (the parameters of $GMM_i^{A_k}$):

$$\begin{cases} w_i^A = \alpha \cdot w_i^{old} + (1-\alpha) \cdot w_i^{new} \\ \mu_i^A = \alpha \cdot \mu_i^{old} + (1-\alpha) \cdot \mu_i^{new} \\ \sigma_i^A = \alpha \cdot (\sigma_i^{old} + (\mu_i^A - \mu_i^{old})(\mu_i^A - \mu_i^{old})^T) + (1-\alpha) \cdot (\sigma_i^{new} + (\mu_i^A - \mu_i^{new})(\mu_i^A - \mu_i^{new})^T) \end{cases} \quad (3)$$

where $\{w_i^A, \mu_i^A, \sigma_i^A\}$ are the weight, mean, and variance of the adapted model for the $i$-th Gaussian in the GMM. $\{w_i^{old}, \mu_i^{old}, \sigma_i^{old}\}$ are the parameters of the initial model $\theta^{Old}$, and $\{w_i^{new}, \mu_i^{new}, \sigma_i^{new}\}$ are the updated parameters from the regular EM algorithm in Step 1. $\alpha$ is the weighting factor to control the balance between the initial model and the new estimates.

## 5. CONFIDENT-FRAME-BASED RECOGNITION ALGORITHM

After the activities are described by the combination of CFV-based GMMs, we can construct a GMM classifier $C_i$ for each CFV $F_i$ with the MAP (Maximum a Posteriori) principle, as shown in Eqn (4):

$$\begin{cases} C_i(F_i(t)) = \arg\max_{A_k} P(A_k | F_i(t)) \\ P(A_k | F_i(t)) = {p(F_i(t) | A_k) \cdot P(A_k)} \Big/ {p(F_i(t))} \end{cases} \quad (4)$$

where $p(F_i(t)|A_k)$ is the likelihood function for the observed CFV of $F_i(t)$ in category $i$ at frame $t$, given activity $A_k$, calculated by Eqn (2). $P(A_k)$ is the probability for activity $A_k$ and $p(F_i(t))$ is the likelihood function for the CFV $F_i(t)$.

The GMM classifiers for different CFVs will differentiate activities with different confidence (e.g., the classifier $C_{CBS}$ is more capable to differentiate *Inactive* and *Fighting*, while the classifier $C_{Speed}$ may have more difficulty in doing so), leading to various possible inconsistencies among results from classifiers for different CFVs. Thus, it is desirable to fuse the classification results from different classifiers to obtain the final improved result. In the following, we propose a Confident Frame-based Recognition algorithm (CFR) to improve the recognition accuracy.

## 5.1 Combining the Global Model and Local Model for Improved Recognition

Due to the uncertain nature of human actions, samples of the same action may be dispersed or clustered into several groups. The 'global' model derived from the whole set of training data collected from a large population of individuals with significant variations may not give good results in classifying activities associated with an individual. In this section, we introduce the idea of using local models to improve the accuracy of recognizing activities.

Using Fig. 4 as an example, there are two global models: $A_k$ for activity *walking* and $A_j$ for activity *running*. The cross samples in the figure are frame positions in the feature space with each cluster of crosses representing one period of action taken by one person. Due to the non-rigidness of human actions, each individual person's activity pattern may be 'far' from the 'normal' patterns of the global model. In this example, if Person 1 walks (cluster *W1* in Fig. 4) faster than normal people and Person 2 walks (cluster *W2* in Fig. 4) slower than normal people, then most of the samples in both clusters will be 'far' from the center of the 'global' model for $A_k$. When using the global model to perform classification on Cluster *W1*, only a few samples in *W1* can be correctly classified. The other samples in *W1* may be misclassified as $A_j$. However, based on the self-correlation of samples within the same period of action, if we use those samples that are well recognized by the global model (boldfaced crosses in Fig. 4) to generate 'local' models, it could greatly help the global model to recognize other samples.

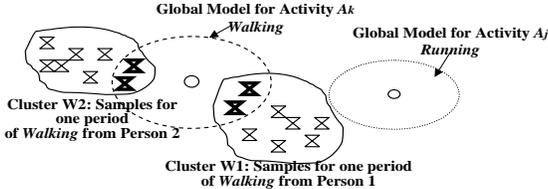

Fig. 4. Global and local models.

Based on the idea described in the above example, our proposed Confident-Frame-based Recognition (CFR) algorithm can be described as follows:

1. *For an activity $A_k$, we use the 'global' model to detect frames $T_k$ which have high confidence for recognizing $A_k$, instead of trying to match every frame directly using the global model. We call $T_k$ Confident Frames,* while the rest of the frames are called *Left Frames* (denoted as $L_k$) as shown in Fig. 5. Many methods can be used to detect *confident frames*, such as weighted average [37][38] or weighted multiplication [36] of recognition results from the CFVs. In this paper, weighted average is used to detect *confident frames*:

$$t = \begin{cases} T_k & if \ \sum_i w_{k,i} \cdot P(A_k | F_i(t)) > th_k \quad (5) \\ L_k & if \ \sum_i w_{k,i} \cdot P(A_k | F_i(t)) \leq th_k \quad (6) \end{cases}$$

where $t$ is the current frame, $T_k$ is a *Confident Frame* and $L_k$ is a *Left Frame*. $P(A_k|F_i(t))$ is the recognition result from the global model of CFV $F_i$. $P(A_k|F_i(t))$ can be calculated by Eqn (4). $w_{k,i}$ is the weight for the global model result of CFV $F_i$ under action $A_k$. $th_k$ (>0) is the threshold for detecting *confident frames* for $A_k$. $w_{k,i}$ ($0 \leq w_{k,i} \leq 1$, $\Sigma_i w_{k,i} = 1$) and $th_k$ can be selected by the five-fold cross validation method [35].

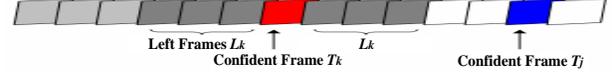

Fig. 5. Confident Frames and Left Frames associated with an activity $A_k$.

2. *The confident frames will be used to generate a 'local' model (or specialized model) for the current period of activity $A_k$. The local model and global model will be used together to classify the Left Frames $L_k$.* The *global model* is used first to select the most possible candidate events, and then the *local model* is used to decide the best one among the candidate events. The decision on the best candidate event is based on our proposed Multi-category Dynamic Partial Function (M-DPF) which is extended from Dynamic Partial Function (DPF) [21]. The M-DPF is described by

$$D_w(X,Y) = k_1 \cdot \left( \sum_{f_{il} \in F_1 \setminus \Delta} w_{il} \cdot (f_{il}^X - f_{il}^Y)^r \right)^{1/r} + \ldots + k_n \cdot \left( \sum_{f_{in} \in F_n \setminus \Delta} w_{in} \cdot (f_{in}^X - f_{in}^Y)^r \right)^{1/r} \quad (7)$$

where X and Y are two feature sets. $F_i \setminus \Delta$ are features in $F_i$ but not in $\Delta$. $w_{ij}$ is the weight for the $i$-th feature $f_{ij}$ in CFV $F_j$. $k_j$ is the weight for CFV of $F_j$. $r$ is a constant parameter.
$\Delta = \{the\ largest\ n\ (f^X - f^Y)'s\ of\ the\ set\ of\ [F_1^X - F_1^Y, F_2^X - F_2^Y, \ldots, F_n^X - F_n^Y]\}$, $F_j = [f_{1j}, f_{2j}, f_{3j}, \ldots f_{mj}]^T$ is the CFV for category $j$. The M-DPF in Eqn (7) is used to measure the dissimilarity between the *confident frame* $T_k$ with the feature set X and *left frame* $L_k$ with the feature set Y (the testing sample). Since frames during an activity $A_k$ represent the *same consistent* action of the *same* person, the self-correlation between the frames during $A_k$ should be higher than the correlation between the frames inside the duration of $A_k$ and the frames outside the duration of $A_k$. This means that normally $L_k$ will be more similar to $T_k$ than $T_j$ if $j \neq k$, as shown in Fig. 5.

## 5.2 Summary of the CFR Process

The Confident-Frame based Recognition process is summarized as follows:

**a.** For a given video sequence, first detect all *confident frames* associated with each activity by Eqn (5).

**b.** For each *Left Frame* $t_L$, pick the two most possible candidate activities for this frame by Eqn (8):

$$\begin{cases} candi1 = \arg\max_k \sum_i w_{k,i} \cdot P(A_k | F_i(t)) \\ candi2 = \arg\max_{k, k \neq candi1} \sum_i w_{k,i} \cdot P(A_k | F_i(t)) \end{cases} \quad (8)$$

**c.** Select the two confident frames $T_{candi1}$ and $T_{candi2}$ corresponding to the two most possible candidate activities which are temporally closest to $t_L$. If we cannot find $T_{candi}$ in the duration of the activity associated with the current object, the temporally closest $T_{candi}$ of a different object with the same activity can be selected. Then calculate the dissimilarity of $D_w(T_{candi1}, t_L)$ and $D_w(T_{candi2}, t_L)$ by Eqn (7).

**d.** The candidate with smaller $D_w$ will be the result for frame $t_L$.

In the above process, the global model and the local model are used together for classifying the *left frames*, in order to increase the accuracy of the recognition. The global model based on the GMMs is first used to select the two most possible candidate activities, then the local model (confident frame-based dissimilarity checking) is used to classify a left frame into one of the two candidate activities.

*5.3 Discussion of CFR and Comparison with Other Methods*

Since the CFR method can be considered as a method for combining the results from the CFV classifiers, it can be compared to other Multi-steam Fusion (MF) methods. Compared with most MF methods [22,35,36,37,38] or other event detection methods [10-17] described in Section 1, the major difference of our proposed CFR algorithm is the introduction of the local models to help recognize the *left frames*. With the introduction of local models, the CFR algorithm has the following four advantages:

**(1)** Most MF methods and other methods focus on detecting the occurrence of events and are normally not good at detecting the boundary between two actions precisely, while our CFR method can effectively detect the starting and ending point of activities.

**(2)** In cases when it fails to detect any confident frame during period $P_{Ak}$ of action $A_k$, CFR may still be able to detect the event by checking the dissimilarity with local models (confident frames) of $A_k$ outside $P_{Ak}$, as in Fig. 6. This makes it more robust and accurate compare to MF methods.

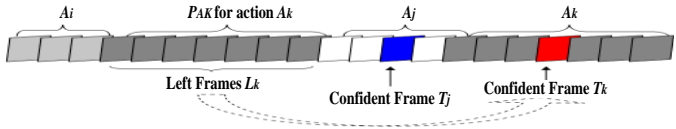

Fig. 6. When failing to detect any confident frame during period $P_{Ak}$, CFR may still be able to detect the event by checking the dissimilarity with confident frames of $A_k$ outside $P_{Ak}$.

**(3)** Many MF methods [35-38] need to carefully select the fusion parameters in order for these methods to perform well on each sample in the test set. This parameter selection will become more difficult when the number of samples or activities increases. However, CFR only requires the parameters to work well with the local model (confident frames) for each activity period, which will greatly facilitate the parameter selection process.

**(4)** By introducing the local model into the activity recognition, we can also take the advantage of using more features. Some kinds of features such as *object location* may not be suitable for differentiating activities for the classifiers. For example, many activities can take place anywhere, therefore *object location* is not able to differentiate them. However, when checking dissimilarities between the *Confident Frames* and the *Left Frames*, these features will be useful. Therefore, CFR enables the inclusion of more features to facilitate the recognition.

## 6. EXPERIMENTAL RESULTS

In this section, we show experimental results for our proposed methods. The experimental results of the *CFR algorithm to improve recognition accuracy* are shown in Section 6.1. In Section 6.2, experimental results are shown to justify the effectiveness of the proposed *LTS event training method*. Section 6.3 shows the results to justify *the flexibility of our algorithm to add new events*.

*6.1 Experimental Results for the CFR algorithm*

We perform experiments using the PETS'04 database [20], and try to recognize five activities: *Inactive, Active, Walking, Running, and Fighting.* The total numbers of video frames for each activity are listed in Table 1.

Table 1
Number of positive and negative samples (video frames) for each activity

|  | # of Positive Samples | # of Negative Samples |
|---|---|---|
| *Inactive* | 9077 | 18276 |
| *Active* | 3236 | 24117 |
| *Walking* | 14144 | 13209 |
| *Running* | 490 | 26863 |
| *Fighting* | 406 | 26947 |

Table 2
CFV and Feature definitions

| CFV | Feature | Definition of f |
|---|---|---|
| $CFV_{bm}=$ [$f_{c\_mb\_sz}$, $f_{c\_mb\_wd}$, $f_{c\_mb\_ht}$] | $f_{c\_mb\_sz}$ | $\sum_{i=t-k}^{t+k} \left| \frac{w_{mbb}^i \cdot h_{mbb}^i - w_{mbb}^{i-1} \cdot h_{mbb}^{i-1}}{w_{mbb}^{i-1} \cdot h_{mbb}^{i-1}} \right| \Big/ 2k+1$ |
|  | $f_{c\_mb\_wd}$ | $\sum_{i=t-k}^{t+k} \left| \frac{w_{mbb}^i - w_{mbb}^{i-1}}{w_{mbb}^{i-1}} \right| \Big/ 2k+1$ |
|  | $f_{c\_mb\_ht}$ | $\sum_{i=t-k}^{t+k} \left| \frac{h_{mbb}^i - h_{mbb}^{i-1}}{h_{mbb}^{i-1}} \right| \Big/ 2k+1$ |
| $CFV_{bt}=$ [$f_{avg\_spd}$, $f_{avg\_vct}$, $f_{mean\_vct}$] | $f_{avg\_spd}$ | $\sum_{i=t-k}^{t+k} \sqrt{(x_{MBB}^i - x_{MBB}^{i-1})^2 + (y_{MBB}^i - y_{MBB}^{i-1})^2} \Big/ 2k+1$ |
|  | $f_{avg\_vct}$ | $\sqrt{\left(\frac{x_{MBB}^{t+k} - x_{MBB}^{t-k}}{2k+1}\right)^2 + \left(\frac{y_{MBB}^{t+k} - y_{MBB}^{t-k}}{2k+1}\right)^2}$ |
|  | $f_{mean\_vct}$ | $\sum_{i=-(k-1)}^{k} \sqrt{\left(\frac{x_{MBB}^{t+k} - x_{MBB}^{t-i}}{k+i+1}\right)^2 + \left(\frac{y_{MBB}^{t+k} - y_{MBB}^{t-i}}{k+i+1}\right)^2} \Big/ 2k$ |

Note: $CFV_{bm}$ is the CFV for body movement, and $CFV_{bt}$ is the CFV for body translation. $c\_mb\_sz$, $c\_mb\_ht$ and $c\_mb\_wd$ represent *change of MBB size, height and width*, respectively. $h_{mbb}$ is the height for object's MBB, $w_{mbb}$ is the width for object's MBB, ($x_{MBB}$, $y_{MBB}$) is the center of MBB. $t$ is the current frame number. $t$-$k$ and $t$+$k$ indicate previous and future $k$ frames relative to the current frame. In our experiment, we set $k$ to 4.

For simplicity, we only use the Minimum Bounding Box (MBB) information (which is the smallest rectangular box that includes the object [11]) to derive *all* the features used for activity recognition. Note that the proposed algorithm is not limited to MBB features. Other more sophisticated features [18-19] can easily be applied to our algorithm to give better results. It should also be noted that in our experiments, some activities do not have enough training data. The definitions of features used for activity recognition are listed in the third column of Table 2. The features are grouped into two CFVs by the K-L distances in Eqn (2), with CFV$_{bm}$ for the category *body movement*, and CFV$_{bt}$ for the category *body translation*. The K-L distances between the features in Table 2 for *one set of training data* are listed in Table 3. The grouping result is shown by the circles in Table 3. The matrix is similar for other training sets and the grouping results are the same.

Table 3
K-L distance for features in Table 2 for one set of training data

|  | $f_{c\_mb\_sz}$ | $f_{c\_mb\_wd}$ | $f_{c\_mb\_ht}$ | $f_{avg\_spd}$ | $f_{avg\_vct}$ | $f_{mean\_vct}$ |
|---|---|---|---|---|---|---|
| $f_{c\_mb\_sz}$ |  | 34.42 | 19.35 | 9463.4 | 6471.5 | 6058.8 |
| $f_{c\_mb\_wd}$ | 34.42 |  | 13.36 | 8683.1 |  | 6239.4 |
| $f_{c\_mb\_ht}$ | 19.35 | 13.36 |  | 4773.1 | 3770.5 | 3382.1 |
| $f_{avg\_spd}$ | 9463.4 | 8683.1 | 4773.1 |  | 11.37 | 12.95 |
| $f_{avg\_vct}$ | 6471.5 | 6941.2 | 3770.5 | 11.37 |  | 11.90 |
| $f_{mean\_vct}$ | 6058.8 | 6239.4 | 3382.1 | 12.95 | 11.90 |  |

(Circles: CFVs are formed by grouping features with small distances)

In order to exclude the effect of a tracking algorithm, we use the ground-truth tracking data which is available in the PETS'04 dataset to get the MBB information. In practice, various practical tracking methods [10,23] can be used to obtain the MBB information. Furthermore, the features in Table 2 are calculated by averaging several consecutive frames to improve the robustness to the possible tracking error.

Due to the inclusion of the local model, more features for the M-DPF dissimilarity checking become useful. The new added features are listed in Table 4. When checking the M-DPF dissimilarity by Eqn (7), we set $r=2$, $k_i=1$, and $w_i=1/\sigma_i$, where $\sigma_i$ is the standard deviation of feature $f_i$. The $w_i$'s for the features of $x^t_{MBB}$, $y^t_{MBB}$, and $d^t_{ob}$ are set to 1. We discard the two features with the largest distances (i.e., $n=2$ in Eqn (7)).

Table 4
New added features for the DPF dissimilarity checking

| Feature Name | Definition |
|---|---|
| $x^t_{MBB}$ | x-axis position of the center of MBB |
| $y^t_{MBB}$ | y-axis position of the center of MBB |
| $d^t_{ob}$ | Duration: # of frames since the object *ob* appears |
| $h_{mbb}$ | height for object's MBB |
| $w_{mbb}$ | width for object's MBB |
| $r^t_{mbb}$ | $r^t_{mbb}=h^t_{mbb}/w^t_{mbb}$ |
| $size^t_{mbb}$ | $size^t_{mbb}=h^t_{mbb} \cdot w^t_{mbb}$ |

*6.1.1 Frame Error Rate Comparison for different methods*

In this experiment, we compare the frame-level error rate of the following five methods for fusing results from multiple streams. Frame-level error rate measures the recognition accuracy for *each* video frame.

**(i) Weighted Average** [37,38] (WA in Table 5). Use a weighted average of results from the two CFVs, as in Eqn (9).

$$t = A_k \quad if \quad \left(\sum_i w_{i,k} \cdot P(A_k | F_i(t))\right) > \max_{j, j \neq k}\left\{\sum_i w_{i,j} \cdot P(A_j | F_i(t))\right\} \quad (9)$$

where *t* is the current frames (or sample). The definition of $w_{i,j}$ and $P(A_j|F_i(t))$ are the same as in Eqn (5).

**(ii) Weighted Multiplication** [35,36] (WM in Table 5). The results for the two classifiers are combined by $p(F_{bm}|A_k)^{wk} \cdot p(F_{bt}|A_k)^{(1-wk)}$, where $p(F_{bm}|A_k)$ and $p(F_{bt}|A_k)$ are GMM distributions for CFV$_{bm}$ and CFV$_{bt}$. $w_k$ is the weight representing the relative reliability of the two CFVs for $A_k$.

**(iii) AVC method** [22] (AVC in Table 5). In [22], the histograms of audio and video features are combined to form an Audio-Video Co-occurrence (AVC) matrix. In our experiment, we create two labeled histograms for the two CFVs for each activity (based on the method in [10]) and use them to replace the histograms of audio and video features in [22]. There will be one AVC matrix for each activity. After the AVC matrix for activity $A_k$ is created, the activity $A_k$ can then be detected based on the AVC matrix.

**(iv) Early Integration** [35] (EI in Table 5). Use *one* GMM model for the whole six features in Table 2.

**(v) The proposed CFR algorithm** (CFR in Table 5). Use the weighted average of GMM as a global model to detect *confident frames and* use them as the local model, and then combine the global and local models to detect the *left frames*.

The experiments are performed under 50% Training and 50% Testing. We perform five independent experiments and average the results. The results are shown in Table 5. In order to show the contribution of each individual CFV, we also include the results of using only CFV$_{bm}$ classifier (C$_{bm}$ in Table 5) or only CFV$_{bt}$ classifier (C$_{bt}$ in Table 5).

In Table 5, the Misdetection *(Miss)* rate and the False Alarm (FA) rate [10] are compared. In the last row of Table 5, we include the Total Frame Error Rate (TFER) which is defined by $N_{t\_miss} / N_{t\_f}$, where $N_{t\_miss}$ is the total number of misdetection frames for all activities and $N_{t\_f}$ is the total number of frames in the test set. TFER reflects the overall performance of each algorithm in recognizing all these five activities.

From Table 5, we can see that the proposed CFR algorithm, which introduces the local model to help detect activities, has the best recognition performance compared with other methods. Furthermore, for activities such as *Active, Running*, and *Fighting* where the GMM classifiers have high *Misdetection rates* (*miss* in Table 5), our CFR algorithm can greatly improve the detection performance.

Table 5
Frame-level Error Rate Results for 50% Training and 50% Testing

|  |  | C$_{bt}$ | C$_{bm}$ | WA | WM | AVC | EI | CFR |
|---|---|---|---|---|---|---|---|---|
| Inactive | Miss (%) | 3.3 | 4.8 | 0.61 | 0.43 | 6.43 | 6.9 | 0.87 |
|  | FA (%) | 3.1 | 2.9 | 6.2 | 6.91 | 0.83 | 1.4 | 1.6 |
| Active | Miss (%) | 40.1 | 56.8 | 46.6 | 43.57 | 20.21 | 25.61 | 12.9 |
|  | FA (%) | 2.72 | 4.06 | 0.51 | 0.76 | 1.64 | 1.87 | 1.09 |
| Walking | Miss (%) | 4.02 | 7.15 | 2.6 | 2.87 | 8.32 | 1.32 | 2.41 |
|  | FA (%) | 11.84 | 19.25 | 16.82 | 16.1 | 3.9 | 5.57 | 5.11 |
| Running | Miss (%) | 56.7 | 90.31 | 77.85 | 83.3 | 44.72 | 52.06 | 26.25 |
|  | FA (%) | 0.2 | 0 | 0.04 | 0.01 | 1.43 | 0.4 | 0.27 |
| Fighting | Miss (%) | 70.2 | 76.53 | 71.41 | 73.16 | 59.57 | 56.62 | 40.81 |
|  | FA (%) | 0.07 | 0.16 | 0.02 | 0.03 | 1.47 | 0.38 | 0.2 |
| **TFER** |  | 12.42 | 15.73 | 11.28 | 11.46 | 10.24 | 8.81 | 4.09 |

*6.1.2 Activity-level Error Rate Comparison*

In the previous section, we showed experimental results for the frame-level error rates. However, in some scenarios, people are more interested in the error rate in the activity level (i.e., the rate of missing an activity when it happens). In these cases, frame-level error rates may not be able to measure the performance accurately. For example, in Fig. 7, the two results have the same frame-level error rates while their activity-level error rates are different (*Recognition Result A* has a lower activity-level error rate because it detects both of the $A_2$ actions while *Recognition Result B* only detects one).

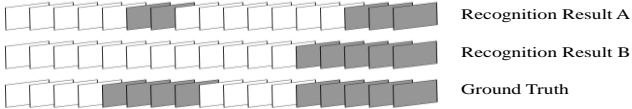

Fig. 7. Recognition results comparison. (White frame sequence: action $A_1$, grey frame sequence: action $A_2$)

In this section, we compare the activity-level error rate performance. First we define the time interval $[t_1, t_2]$ to be an *Activity Clip* of activity $A_k$ if:

$$\begin{cases} L_{grd}(t_1) \neq L(t_1-1) & (\text{not } A_k \text{ before } t_1) \\ L_{grd}(t) = A_k, \forall t \in [t_1, t_2] & (\text{all label is A during } [t_1,t_2]) \\ L_{grd}(t_2) \neq L(t_2+1) & (\text{not } A_k \text{ after } t_2) \end{cases}$$

where $L_{grd}(t)$ is the ground-truth activity label of frame $t$.

The Activity-level Error Rate (AER) in this experiment is then defined as $AER=N_{k,miss}/N_{k,total}$, where $N_{k,miss}$ is the total number of *missed Activity Clip* in Eqn (10) for activity $A_k$. $N_{k,total}$ is the total number of *Activity Clips* for $A_k$. An *Activity Clip* of time interval $[t_1, t_2]$ is a *missed Activity Clip* if:

$$L_{grd}(t) \neq L_{recognized}(t), \forall t \in [t_1, t_2] \quad (10)$$

where $L_{grd}(t)$ is the ground-truth activity label at frame $t$. $L_{recognized}(t)$ is the recognition result at frame $t$.

In Table 6, we compared the AER performance of the five methods described in Section 6.1.1.

Table 6
Activity-level Error Rate Results for 50% Training and 50% Testing

|  | $C_{bt}$ | $C_{bm}$ | WA | WM | AVC | EI | CFR |
|---|---|---|---|---|---|---|---|
| Inactive | 1.8% | 2.04% | 0.6% | 0.6% | 3.6% | 1.2% | 0.6% |
| Active | 17.67% | 20.3% | 12.67% | 13.33% | 15.5% | 12.67% | 9.56% |
| Walking | 3.77% | 5.68% | 3.4% | 4.32% | 6.16% | 3.86% | 2.31% |
| Running | 41.33% | 63.67% | 36.33% | 43% | 27.33% | 36.33% | 24% |
| Fighting | 50% | 50% | 50% | 50% | 46.67% | 50% | 36.67% |

Some important observations from Table 6 are listed below.

**(1)** Compared to the *Frame-level Misdetection (Miss) Rates* in Table 5, some methods have much closer performances in *Activity-level Error Rates (AERs)* (e.g., the *Miss* rate for *running* of EI in Table 5 is more than 25% lower than that of WA, however, their *AERs* are the same in Table 6). This is because these two rates (*Miss* and *AER*) reflect different aspects of the recognition methods. The *Miss* rate reflects more on the ability of the methods to precisely locate the boundary of events (i.e., the ability to recognize all frames between the starting and ending points of the events), while the AER reflects more on the ability of the methods to detect events when they happen (i.e., the ability to detect at least one frame when the event occurs).

Comparing Table 5 and Table 6, we find that most methods have a much lower *AER* than the *Miss* rate (especially for activities with high *Miss* rates such as *active*, *running* and *fighting*). This means that most of these methods are more capable of detecting the existence of the activities than *precisely* locating their boundaries. Compared to these methods, the proposed CFR algorithm has a similar *AER* but a greatly improved *Miss* rate. This shows that CFR can locate the activity boundaries more precisely.

**(2)** The CFR uses Weighted Average (WA) to detect the *confident frames* as the local model. This means that if an *activity clip* is *missed* by WA, CFR will also fail to detect any *confident frames* in the same *activity clip*. However, the result in Table 6 shows that many of the AERs of CFR are lower than those of WA. This is because when WA fails to detect any *confident frame* during an *activity clip* of $A_k$, CFR may still be able to detect the event by checking the dissimilarity with local models of $A_k$ outside the clip.

**(3)** Based on the previous two observations, we see that the introduction of the *local model* in CFR has two effects: (a) it helps detect the *left frames* within its own *activity clip*, thus locating the clip boundary more precisely and also reducing the frame-level error rates (*Miss* and *FA*), (b) it helps detect other *activity clips* where no confident frame is detected, thus reducing the misdetection rate for activity clips.

**(4)** From Table 6, we can see that the *AERs* of CFR for most activities are close to those of WA. This means that the *AER* performance of CFR mainly depends on the algorithm to detect *confident frames*. Therefore, a suitable confident-frame detection method is important. In this paper, WA is used for detecting confident frames. However, other methods such as WM and AVC can also be applied if they have better performance.

*6.1.3 Experimental Results for Weights and Thresholds Selection*

In several methods such as WA and WM, we need to select a suitable weight (i.e., the $w_{i,k}$ in Eqn (9)) to fuse the results from two CFVs. Furthermore, since the CFR algorithm uses WA to detect the *confident frames*, the weights and thresholds (i.e., $w_{i,k}$ and $th_{i,k}$ in Eqn (5)) also need to be selected for *confident frame* detection. In the previous experiments, all these weights and thresholds are selected by the *five-fold cross validation* method [35]. However, the cross-validation is relatively complicated. We need to try all the possible combinations of parameters. Furthermore, the complexity of the cross-validation algorithm will increase exponentially with the increasing number of parameters. As mentioned in Section 5.3, our proposed CFR algorithm is more robust to the change of weight values since the weights only need to work well on *confident frames* rather than *the whole testing data*. This implies that with the proposed CFR algorithm, we may be able to use a rough weight or use a simpler way to select the parameters. In the following, we show two experimental results to justify this claim.

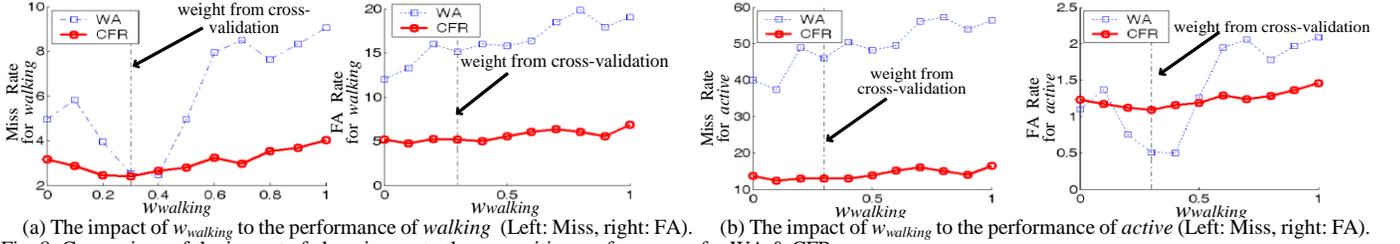

(a) The impact of $w_{walking}$ to the performance of *walking* (Left: Miss, right: FA). (b) The impact of $w_{walking}$ to the performance of *active* (Left: Miss, right: FA).
Fig. 8. Comparison of the impact of changing $w_k$ to the recognition performances for WA & CFR.

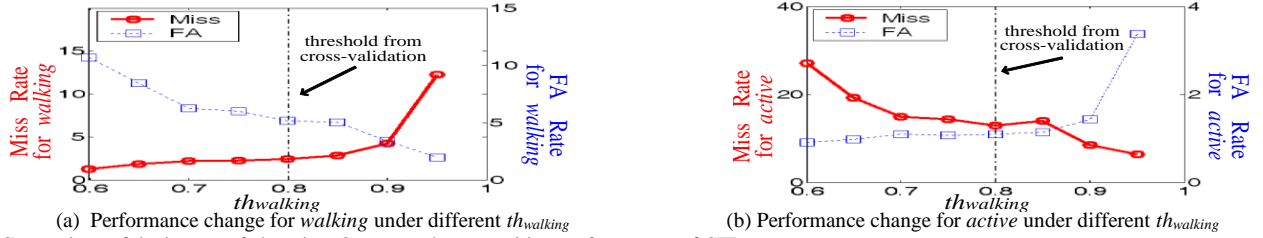

(a) Performance change for *walking* under different $th_{walking}$ (b) Performance change for *active* under different $th_{walking}$
Fig. 9. Comparison of the impact of changing $th_{walking}$ to the recognition performances of CFR.

*6.1.3.1 Experiment 1 for parameter selection*

In this experiment, we justify our claim that the recognition performance of our CFR algorithm is *robust* to the change of parameters. Since the CFR in this paper uses the same method as WA to detect confident frames, we will focus on the comparison of these two methods. Furthermore, since we only have two CFVs in the experiment, the CFR and WA algorithm in Eqn (5) and Eqn (9) can be *re-written* as in Eqn (11) and Eqn (12), respectively.

CFR: $t = T_k \quad if \ (w_k \cdot P(A_k | F_{bm}) + (1-w_k) \cdot P(A_k | F_{bt})) > th_k$ (11)

WA: $t = A_k \quad if \ (w_k \cdot P(A_k | F_{bm}) + (1-w_k) \cdot P(A_k | F_{bt}) > \arg\max_{A_i, A_i \neq A_k}\{w_i \cdot P(A_i | F_{bm}) + (1-w_i) \cdot P(A_i | F_{bt})\}$ (12)

where $F_{bm}$ and $F_{bt}$ represent the features for $CFV_{bm}$ and $CFV_{bt}$, respectively. The definition of $w_k$, $t$, $A_k$, $T_k$ and $th_k$ are the same as in Eqn (5) and Eqn (6).

We first select the parameters ($w_k$ and $th_k$ in Eqn (11) and (12)) by cross-validation. The parameter values selected from the validation set is defined as $w_k^{Valid}$ and $th_k^{Valid}$. Then, we change the weight value $w_k$ for one activity $A_k$ and keep the weight value for other activities unchanged. For the CFR algorithm, we also keep the threshold $th_k$ for *all* activities unchanged. We then use the changed parameter set to perform recognition on the testing data and plot the recognition performance changes.

Fig. 8 (a)-(b) shows the recognition performance (*Miss* and *FA*) change for activities under different $w_k$ values of activity *walking* (i.e., $w_k=w_{walking}$). It is the result from one experiment of 50% Training and 50% Testing. The results from other experiments are similar. Fig. 8 (a) shows the impact of changing $w_{walking}$ to the recognition performance for *walking*, and Fig. 8 (b) shows the impact of $w_{walking}$ to *active*. Fig. 8 shows results when $w_{walking}$ is changed. Similar observations can be found when the weights of other activities are changed.

From Fig. 8 (a)-(b), we can see that the performance of the WA method fluctuates substantially with the change of $w_k$. This reflects that the recognition performance of WA is very sensitive to the change of $w_k$. On the contrary, the recognition performances of our CFR algorithm are quite stable with the change of $w_k$. The performance of CFR is close to those under $w_k^{Valid}$ even when $w_k$ is far from $w_k^{Valid}$ (the dashed vertical line). This justifies that CFR is *robust* to the change of $w_k$.

Since CFR also uses threshold $th_k$ to detect confident frames, a similar experiment is performed on $th_k$ to see its impact on the recognition performances of CFR. We fix all $w_k$'s to be $w_k^{Valid}$. Then we change the value of $th_k$ for activity $A_k$ and keep the threshold value for other activities unchanged. The recognition performances under different $th_k$ values of *walking* (i.e., $th_k=th_{walking}$) are plotted in Fig. 9 (a)-(b). Fig. 9 (a) shows the impact of changing $th_{walking}$ to the performance of *walking*. Fig. 9 (b) shows the impact of $th_{walking}$ to the performance of *active*. Three observations from Fig. 9 are listed below:

**(1)** The performance of CFR is stable when $th_k$ changes within a certain range around $th_k^{Valid}$ (the vertical lines in Fig. 9). This means that *CFR is also robust to the change of $th_k$ within a certain range around $th_k^{Valid}$*. We call this range *stable range*.

**(2)** *A too small or too large value of the threshold $th_k$ will obviously decrease the recognition performance of CFR*. A too small threshold value may include many *false alarm samples* as *confident frames* (an extreme case: if $th_k=0$, it will be exactly the same as the WA method). On the other hand, a too large threshold value may reject most of the samples, making the recognition difficult (an extreme case: if $th_k=1$, there will be almost no *confident frames* detected).

**(3)** Different activities $A_k$ may have different $th_k^{Valid}$. However, since each $th_k$ has a *stable range* around $th_k^{Valid}$, we may still be able to find a *common stable range* for all activities. Our experiments imply that values between 0.65 and 0.8 may be a suitable choice of thresholds for most activities.

The results from Fig. 8 and Fig. 9 justify that our CFR algorithm is robust to the change of parameters $w_k$ and $th_k$. This advantage implies that for the CFR algorithm, we may be able to set the parameters to rough specific values or by a simplified parameter-selection method such as increasing the searching step-size or decreasing the searching range, instead of using the complicated cross-validation method to select the parameters. This is further justified in the following experiment.

*6.1.3.2 Experiment 2 for parameter selection*

In this experiment, we set the $w_k$ for all activities to be 0.5 and $th_k$ for all activities to be 0.7. And then use this parameter set to recognize the activities. We perform five experiments with 50% training and 50% testing and average the result (the same setting as Table 5). The experimental results are listed in Table 7. In order to compare with the results under *cross-validation* parameters, we attach the results of Table 5 (the grey columns).

Table 7
Results under roughly selected parameters

|  |  | $WA_{C-V}$ | $WA_R$ | $WA_{C-V}$ | $WM_R$ | $CFR_{C-V}$ | $CFR_R$ |
|---|---|---|---|---|---|---|---|
| Inact-ive | Miss | 0.61% | 3.82% | 0.43% | 2.58% | 0.87% | 0.94% |
|  | FA | 6.2% | 4.71% | 6.91% | 5.16% | 1.6% | 1.76% |
| Act-ive | Miss | 46.6% | 57.24% | 43.57% | 50.03% | 12.9% | 15.23% |
|  | FA | 0.51% | 2.25% | 0.76% | 3.41% | 1.09% | 1.57% |
| Walk-ing | Miss | 2.6% | 5.29% | 2.87% | 6.1% | 2.41% | 2.17% |
|  | FA | 16.82% | 19.3% | 16.1% | 18.22% | 5.11% | 5.69% |
| Run-ning | Miss | 77.85% | 88.11% | 83.3% | 92.53% | 26.25% | 30.08% |
|  | FA | 0.04% | 0.17% | 0.01% | 0.11% | 0.27% | 0.23% |
| Fight-ing | Miss | 71.41% | 82.59% | 73.16% | 80.4% | 40.81% | 45.1% |
|  | FA | 0.02% | 0.15% | 0.03% | 0.22% | 0.2% | 0.19% |
| TFER |  | 11.28% | 15.48% | 11.46% | 15.53% | 4.09% | 4.62% |

(The gray columns named '*C-V*' are results under cross-validation parameters, the white columns named '*R*' are results under roughly set parameters)

In Table 7, three methods are compared (WA, WM, and CFR). From Table 7, we can see that our proposed CFR algorithm can still perform well under the roughly selected parameters while the performances of both WA and WM methods decrease significantly under this situation. This validates that the CFR algorithm allows us to select parameters through more simplified methods with small impact on the performance. As we will see in Section 6.3, this advantage also increases the flexibility of our algorithm for adding new events.

*6.2 Experimental Results for Training LTS Events*

From Table 5, we can see that the misdetection rate (*Miss*) for activities such *Running* and *Fighting* are relatively high (although our CFR algorithm has significantly improved the *mis*detection rate from other methods). This is because the number of training samples in Table 1 is small. The training samples are not sufficient to model the whole distribution of these activities, which reduces the prediction capability of these models for the unknown data.

We use our proposed *LTS event training* method to deal with the insufficient training data problem, where we adapt both CFVs' GMM models of *Running* from *Walking*, while both CFV GMM models of *Fighting* are adapted from *Active* (which is different from Fig. 3 because *running* itself is also lacking training data). The recognition results based on our adapted-GMM models are shown in Table 8.

The results in Table 8 show the effectiveness of our proposed method in dealing with insufficient training data. We can see that although improved by our algorithm, the misdetection rate for *fighting* is still relatively high. The main reason for this is that besides lacking training data, the features we use (in Table 2) are relatively simple (all from MBB), while the feature distributions of these activities are similar to other activities, making the classification difficult. In order to improve the performance further for the activities, more sophisticated features can be used, or the interaction between different objects can be considered, which will be our future work.

Table 8
Proposed method in dealing with insufficient data

|  |  | $C_{bt\_Dt}$ | $C_{bt\_Adt}$ | $C_{bm\_Dt}$ | $C_{bm\_Adt}$ | $CFR_{Dt}$ | $CFR_{Adt}$ |
|---|---|---|---|---|---|---|---|
| Run-ning | Miss | 56.7% | 37.86% | 90.31% | 70.72% | 26.25% | 14.32% |
|  | FA | 0.2% | 0.21% | 0% | 0.28% | 0.27% | 0.32% |
| Fight-ing | Miss | 70.2% | 64.52% | 76.53% | 68.3% | 40.81% | 29.61% |
|  | FA | 0.07% | 0.1% | 0.16% | 0.15% | 0.2% | 0.22% |

(The gray columns labeled as '*Dt*' are results whose models are modeled directly from the training data, the white columns labeled as '*Adt*' are results whose models are adapted by our proposed method)

*6.3 Experiment Results for the Flexibility of Adding New Events*

We give an example to illustrate the flexibility of adding a new event. In this example, a CFV-based system with two CFVs defined by Table 2 has been trained to detect five activities: *Inactive*, *Active*, *Walking*, *Running*, and *Fighting*. We define a **new event** "*picking up or leaving a bag*". Since there is no ground-truth label for *picking up or leaving a bag* in the dataset, we label it manually. The total number of positive samples for "*picking up or leaving a bag*" is *366*. Note that these samples have been excluded from the dataset in the previous experiments so that they are *new* to the system when the event is added.

As mentioned in Section 3, when new events are added to the system, the existing CFVs may not be enough to differentiate all activities, necessitating the adding of new CFVs. In this example, we *assume* that the two existing CFVs in Table 2 are not enough for differentiating the new "*picking up or leaving a bag*" event. Therefore, we add one more CFV named $CFV_{Change\_of\_Body\_Ratio}$. In $CFV_{Change\_of\_Body\_Ratio}$, there is only one feature $f_{c\_mbb\_ratio}$ which represents *the change of MBB ratio*. The new CFV is defined as:

$$\begin{cases} CFV_{Change\_of\_Body\_Ratio} = [f_{c\_mbb\_ratio}] \\ f_{c\_mbb\_ratio} = \sum_{i=t-k}^{t+k} \left| \frac{\frac{w_{mmb}^i}{h_{mmb}^i} - \frac{w_{mmb}^{i-1}}{h_{mmb}^{i-1}}}{\frac{w_{mmb}^{i-1}}{h_{mmb}^{i-1}}} \right| \Big/ 2k+1 \end{cases}$$

where $k$, $w_{mmb}^i$ and $h_{mmb}^i$ are the same as in Table 2.

Then, the *flexibility* of our algorithm for adding new events *in this example* can be described in the following two points:
**(1)** When the new event *picking up or leaving a bag* was added to the system, we do not need to change or retrain the $CFV_{bm}$ and $CFV_{bt}$ models for events *inactive*, *active*, *walking*, *running* and *fighting*. We only need to train the $CFV_{Change\_of\_Body\_Ratio}$ models for these events as well as all the three CFV models for the new event *picking up or leaving a bag*. The models that need training (white circles) and models that don't need training (grey circles) in this example are shown in Fig. 10. In practical situations, the number of models that do not need training is much larger than the number of models that need training.

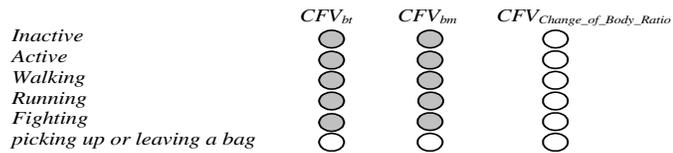

Fig. 10. Models need training or do not need training. (Grey circles: models do not need training, White circles: Models need training)

**(2)** Since the new $CFV_{Change\_of\_Body\_Ratio}$ is added for each event, we need to update parameters that fuse these CFV models ($w_{i,k}$ and $th_k$ in Eqn (5)). However, as mentioned, the CFR algorithm is *robust* to the change of these parameters. This means that we can set these parameters roughly or by a simple parameter selection method, instead of performing the complicated cross-validation method to update the parameters.

Based on the above two points, in the experiment, we can train the new system through a simple way by (1) only training the white labeled CFV models in Fig. 10, and (2) setting the weights and thresholds roughly (here we set all weights to be 0.33 and all thresholds to be 0.7). Table 9 (white column) shows the results for 50% training and 50% testing (the setting is the same as in Table 5). Table 9 (grey column) shows the recognition results under cross-validation parameters.

Table 9
Experimental results for adding new event

|  |  | $CFR_{C-V}$ | $CFR_R$ |
|---|---|---|---|
| Inactive | Miss | 0.9% | 0.95% |
|  | FA | 1.43% | 1.88% |
| Active | Miss | 16.26% | 18.52% |
|  | FA | 1.02% | 1.6% |
| Walking | Miss | 3.07% | 2.86% |
|  | FA | 5.2% | 5.96% |
| Running | Miss | 23.94% | 27.8% |
|  | FA | 0.31% | 0.28% |
| Fighting | Miss | 46.31% | 51.55% |
|  | FA | 0.16% | 0.11% |
| Picking up or leaving a bag | Miss | 36.94% | 40.5% |
|  | FA | 0.19% | 0.17% |
| TFER |  | 4.55% | 5.12% |

(The gray columns named '*C-V*' are results under cross-validation parameters, the white columns named '*R*' are results under roughly set parameters)

From Table 9, we can see that when the system is adapted to include the new event through a simple manner by our algorithm, we still can achieve good results close to those under cross-validation parameters. This justifies the flexibility of the algorithm.

## 7. CONCLUSION

In this paper, we made the following three contributions: (1) We propose to represent activities by the combination of CFV-based models which has good flexibility in representing activities as well as in *handling new events*. (2) Based on the CFV-based representation, we propose a method to deal with the *model training problems for events which lack training data (LTS events)*. (3) We also propose a Confident-Frame based recognition algorithm which is capable of *improving the recognition accuracy*. Experimental results demonstrate the effectiveness of our proposed methods.